%% file: preprint.tex
\documentclass[11pt]{article}

\usepackage[preprint]{acl}
\usepackage[safe]{tipa}

\usepackage{times}
\usepackage{latexsym}

\usepackage[T1]{fontenc}
\usepackage[utf8]{inputenc}

\usepackage{microtype}

\usepackage{inconsolata}

\usepackage{graphicx}
\usepackage{tabularx}

\usepackage{xcolor}
\usepackage{multirow}
\usepackage{booktabs}
\usepackage{amsmath}
\usepackage{fancyvrb}
\usepackage[T1]{fontenc}
\usepackage{tipa}
\usepackage[utf8]{inputenc}
\usepackage{newunicodechar}
\usepackage{float}

\newunicodechar{́}{\'{}}      % U+0301 Combining acute accent
\newunicodechar{̇}{\.{}}      % U+0307 Combining dot above
\newunicodechar{̨}{\k{}}      % U+0328 Combining ogonek
\newunicodechar{̌}{\v{}}      % U+030C Combining caron / háček

\newunicodechar{ȟ}{\v{h}}    % U+021F h with caron
\newunicodechar{š}{\v{s}}    % U+0161 s with caron
\newunicodechar{ṡ}{\.{s}}    % U+1E61 s with dot above
\newunicodechar{ḣ}{\.{h}}    % U+1E23 h with dot above
\newunicodechar{ą}{\k{a}}    % U+0105 a with ogonek
\newunicodechar{ŋ}{\textipa{N}} % U+014B eng using TIPA
\newunicodechar{‘}{`}        % U+2018 Left single quote
\newunicodechar{’}{'}        % U+2019 Right single quote

\usepackage[nameinlink,capitalize,noabbrev]{cleveref}

\crefname{equation}{equation}{equations}   % "equation" is lowercased, overriding capitalize option
\crefname{footnote}{footnote}{footnotes}   % "footnote" is lowercased, overriding capitalize option
\crefname{section}{\S}{\S\S}
\Crefname{section}{\S}{\S\S}    % must define start-of-sentence version explicitly since \S isn't a letter
\crefformat{section}{#2\S#1#3}  % remove space between section symbol and the number, and include \S in the hyperlink
\Crefformat{section}{#2\S#1#3}
\crefrangeformat{section}{\S\S#3#1#4--#5#2#6}
\Crefrangeformat{section}{\S\S#3#1#4--#5#2#6}
\crefformat{section}{#2\S#1#3}  % remove space between section symbol and the number, and include \S in the hyperlink
\crefrangeformat{section}{\S\S#3#1#4--#5#2#6}

\title{Design Considerations in Low-Resource Archival OCR: A Case Study in Lakota}

\author{
  \textbf{Téa Wright}$^1$ \quad 
  \textbf{Laura X. Ma}$^1$ \quad 
  \textbf{Claudia Iron Hawk}$^1$ \quad 
  \textbf{Hannah J. Haynie}$^2$ \\
  \textbf{Alexis Palmer}$^2$ \quad 
  \textbf{Alane Suhr}$^1$ \quad 
  \textbf{Abteen Ebrahimi}$^2$ \quad 
  \textbf{Katharina von der Wense}$^2$ \\[0.5em]
  $^1$University of California, Berkeley \quad\quad 
  $^2$University of Colorado, Boulder \\
  {\small \texttt{teaywright@berkeley.edu}}
}
\begin{document}
\maketitle
\begin{abstract}
\input{sections/abstract}
\end{abstract}

\input{sections/intro}
\input{sections/related}
\input{sections/lakota_and_data}
\input{sections/ocr}

\input{sections/analysis.tex}

\input{sections/conclusion}
\input{sections/ethics}

\input{sections/limitations}
\input{sections/acknowledgements}

\bibliography{custom}

\appendix

\input{sections/appendix/lakota}
\input{sections/appendix/examples}
\input{sections/appendix/implementation_details}
\input{sections/appendix/full_results}
\input{sections/appendix/additional_analyses}

\end{document}

%% file: sections/abstract.tex
  Building and expanding access to digital archives supports language reclamation and creates lasting resources for community revitalization. Designing effective  tools to support these efforts requires taking into account the community-, language-, and document-specific context. We investigate how small ($\leq$3B) vision-language models (VLMs) can be adapted as optical character recognition (OCR) tools for Lakota, a North American Indigenous language. In addition to benchmarking modeling approaches, we inspect four practical facets of low-resource archival OCR that arise in previous work and conversations with community members: constrained training budgets, segmentation noise, scan degradation, and orthographic variance. We demonstrate that fine-tuning VLMs for OCR post-correction on 35 pages of transcribed data decreases character error rate (CER) from 17.1 to 3.9\% on a bilingual corpus, and find that training only with nine pages increases CER by only 1.1\% over this. 
 However, evaluating under realistic automatic segmentation nearly doubles the CER of the best-performing models, stressing the importance of taking realistic use cases into account. We hope this work encourages future research on developing language technologies grounded in the workflows and constraints of ongoing revitalization efforts. 

%% file: sections/intro.tex
\section{Introduction}
\input{figs/fig1}
Programs of language revitalization and reclamation are greatly supported by the archival curation of language resources. These efforts can enable community members to easily access cultural knowledge or provide educators with resources to create learning materials. 

For languages with available written materials, the process of transcribing documents for digitization can be a manual and slow undertaking. These projects can be further facilitated by the automatic digitization of materials through optical character recognition (OCR). 
However, languages undergoing processes of revitalization are often under-served by the development of such technologies, which depends on access to large amounts of in-domain data---for example, for training neural models. 
This results in technologies that are largely useless out-of-the-box, primarily because they were designed with high-resource languages in mind.
To build tools that serve efforts in language revitalization, the development of language technologies must take into account the needs and constraints of language communities that break the assumptions upon which such technologies have been predominantly built.

Technologies for Indigenous languages frequently fail to align with real-world deployment settings \citep{rice2025interdisciplinary}, producing de-contextualized models rather than tools that serve community needs \citep{cooper2024s}. For example, in low-resource OCR, existing work often relies on unrealistic assumptions like line-level image crops annotated by hand \citep{Agarwal2024-au}. Beyond optimizing for evaluation metrics on static data, models must be built to actually function in downstream workflows \citep{gessler-etal-2026-missing}. Doing so requires addressing practical constraints of community-facing digitization including scarce transcribed data, orthographic variance across document collections, and faded historical documents. 

In this work, we conduct a case study in digitizing Lakota documents in collaboration with community members involved in language revitalization work. We frame our study as a two-part exploration: (a) evaluating traditional and VLM-based OCR systems through benchmarking on a newly transcribed dataset and (b) exploring the effects of realistic settings informed by community needs and constraints; this includes training budget, segmentation strategies, scan quality, and orthographic variance. \cref{fig:pipeline} highlights the OCR pipeline as well as where data- or language- specific considerations may arise.

Our findings demonstrate that VLMs fine-tuned both directly for OCR and post-correction are surprisingly sample efficient and can generalize across orthographies. Nonetheless, automatic segmentation remains an issue, nearly doubling the error rates seen on gold line crops. Ultimately, we hope this work encourages academic researchers to ground under-served language technology creation in realistic conditions and align with ongoing community efforts.

%% file: figs/fig1.tex
\begin{figure*}[t]
  \centering
  \includegraphics[width=\textwidth]{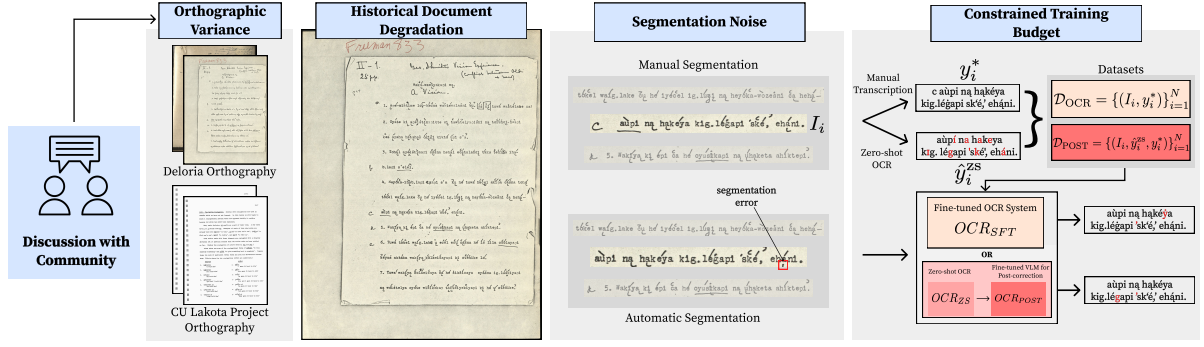}
  \caption{Overview of some of the challenges in Lakota document transcription (e.g., orthographic variation, document degradation, manual vs. automatic segmentation noise, and constrained training budgets) and where they may arise along an OCR training and inference pipeline. }
  \label{fig:pipeline}
\end{figure*}

%% file: sections/related.tex
\section{Related Work}

\paragraph{OCR Methods for Archival and Low-Resource Texts}
Off-the-shelf OCR engines like Google Cloud Vision \citep{fujii2017sequence} and Tesseract \citep{smith2007overview} degrade when applied to low-resource languages \citep{Agarwal2024-au}. To improve transcription without training traditional OCR models from scratch, prior work has relied on text-to-text post-correction, framing the task as sequence-to-sequence translation \citep{Rijhwani2020-aw, Rijhwani2021-sp, lofgren2024post} or prompting generative LLMs \citep{Thomas2024-vd}. Recent work has explored VLM-based OCR, evaluating zero-shot VLMs on diverse scripts \citep{gagnier2026kazakhocr}, fine-tuning VLMs directly \citep{chung2025fine}, or conditioning them on both the noisy OCR text and the image crop \citep{Greif2025-sb}. However, training-based methods for multimodal post-correction remain underexplored.

\paragraph{Language, Data, and Community-Specific Considerations in OCR}
There are several position and qualitative papers that stress the importance of including language communities in the process of creating technologies to ensure artifacts are actually useful \citep{rice2025interdisciplinary, gessler-etal-2026-missing, Mager2023-tu}. Standard OCR evaluations typically benchmark on synthetic data \citep{sasagawa2026synth} or clean scans using gold-standard segmentations \citep{Agarwal2024-au}.  \citet{agarwal2025ailla} develop a post-correction dataset for both structural and textual errors, focusing on how recognition errors are compounded by automated segmentation failures. \citet{gardella2026low} demonstrate that VLMs can produce hallucinations that suppress character error rates. Other work focuses of restoring degraded historical documents for application in digitizing historical archives \citep{Guan2025-xe}.

\paragraph{NLP for Lakota} Existing work in NLP for Lakota remains sparse. Early efforts focused on rule-based linguistic tools, such as \citet{curtis-2014-finite}, who built a morphological analyzer to break down complex Lakota verb structures and potentially support machine translation. More recently, \citet{Parankusham2025-gw} introduced LakotaBERT, the only large language model for Lakota, trained with text extracted via OCR; however, the authors highlight noisy OCR outputs as a limitation. Beyond monolingual models, \citet{robertson2026evaluating} determine that even frontier models cannot translate reliably in Lakota. 

%% file: sections/lakota_and_data.tex
\section{Language Context, Community, and Data}
Lak\v{h}\'{o}tiyapi (also known as Lakota or Lakȟóta) is an endangered Siouan language spoken mainly in North and South Dakota, forming part of the O\v{c}h\'{e}thi \v{S}ak\'{o}wi\textipa{N} language group alongside mutually intelligible Dakota varieties.\footnote{For simplicity, we refer to both Lakota and Dakota varieties collectively as Lakota going forward.} Current speaker numbers are around 2,000, most of whom are elders, though revitalization programs are actively working with L2 learners.  
Lakota also has no single standardized orthography; across more than a century of materials written by linguists and community members, more than ten different orthographies emerged, each using distinct diacritics for phonemes like nasal vowels. We provide more information on the Lakota language in \cref{sec:appendix_lakota}.

\subsection{Revitalization and Digitization Efforts}
Several community-driven initiatives have worked to revitalize the Lakota and Dakota languages. At the Pine Ridge Reservation, Maȟpíya Lúta (formerly Red Cloud Indian School) developed the Lakota Language Project to integrate Lakota into its K--12 curriculum starting in 2008 \citep{Fredericks2018-qz}. Around the same time, at the Standing Rock Reservation, educators built immersion programs and community curriculum efforts at the Lakota Summer Institute and Sitting Bull College \citep{Hauff2019-rf, Hauff2020-wj, Hauff2024-io}. In the digital space, platforms like Ow{\'o}ksape have made learning materials more accessible online \citep{Ullrich2020-cv}.\footnote{We discuss data sovereignty and ethics in \cref{sec:ethics}.}
% \as{I might reference Wóoyake earlier than this if we end up focusing the intro on archival tools}

While recent language technologies for under-served languages have focused on downstream tasks such as interlinear glossing \citep{ginn2024glosslm} or translation \citep{robertson2026evaluating}, we focus on OCR. This decision follows discussions among the authors and with community members working on Lakota revitalization and education. We find that transcription remains a bottleneck in digitization and transcription workflows for community repositories like Wóoyake \citep{bannister2027wooyake}.\footnote{\url{wooyake.org}} Moreover, while automated transcription saves time, in the context of Lakota education, manually performing tasks like glossing and translation can provide meaningful practice for language learners. This work focuses strictly on assessing methods for the initial transcription step.

To digitize a document, practitioners currently rely on proprietary tools like Transkribus to generate an initial transcript. They then manually correct both the produces bounding regions and any OCR errors before moving on to transliteration and glossing. To minimize the friction in this process, we systematically evaluate the viability of OCR models and paradigms on Lakota documents, evaluating in comparison to one another, but also to the proprietary tools currently used.

\subsection{Data}
\label{sec:data}
To evaluate OCR and VLMs on Lakota texts, we constructed a gold-transcribed benchmark. While our broader preservation workflow involved scanning 1,137
historical manuscript pages using a Bookeye 4 overhead scanner, producing character-level ground truth requires manual transcription, which is time-consuming. Our core evaluation dataset therefore consists of 44 representative pages manually transcribed by the authors (see annotation details in \cref{appx:transcription}), sampled across two distinct sources:
\paragraph{Lakota Learning Materials (14 pages):} These pages are samples from four documents from the Lakota Learning Materials and an Elementary Dictionary, which were previously scanned. The texts are cleanly typeset using the Colorado University Lakota Project (CULP) orthography \citep{rood1976beginning}.
\paragraph{Deloria (30 pages):} These pages are sampled from the set of documents scanned by authors, which were written by Lakota linguist Ella Deloria in the 1930's in her respective Orthography \citep{deloria2006dakota}. These manuscripts contain traditional stories, word-by-word literal translations, and natural English translations. The originals are typewritten, and diacritics were written in by hand afterward. They additionally contain notes that explain additional information such as uncommon vocabulary, cultural practices, idioms, and context from consultants. 

A summary of the data is provided in \cref{tab:data_makeup}. In general, the Deloria orthography is significantly more diacritic-heavy than CULP, with Lakota words containing 3.16 times as many diacritics on average. In combination with aged documents and mixed typewritten and handwritten text, the Deloria subset presents a much more challenging OCR problem. Both document sets contain a mix of Lakota and English text. We manually tag each page and bounding box as Lakota, English, or Mixed. Because Mixed text often interleaves English explanations with Lakota terms, we use langdetect\footnote{\url{https://pypi.org/project/langdetect/}} on the gold transcriptions to estimate word-level language distributions. This classifies 70.9\% of words in Mixed text as English (leaving 29.1\% as Lakota). On boxes manually tagged as monolingual Lakota or English, langdetect achieves 90.6\% word-level accuracy.

In line with the CARE Principles for Indigenous Data Governance \citep{carroll2020care} and discussions with community partners, document images and full transcriptions will not be publicly released beyond brief illustrative examples of the data, which can be found in \cref{appx:examples}.

\input{figs/data_makeup}

%% file: figs/data_makeup.tex
\begin{table}[t]
  \centering
  \small
  \setlength{\tabcolsep}{6pt} % Adjust horizontal column padding if needed
  \begin{tabular}{l rrrr}
    \toprule
    \textbf{Sub-corpus} & \textbf{Pages} & \textbf{Lines} & \textbf{Chars} & \textbf{Lakota (\%)} \\
    \midrule
    \multicolumn{5}{l}{\textit{Deloria}} \\
    \addlinespace[2pt]
    \hspace{1em}Lakota & 14 & 299 & 20,655 & 96.5\% \\
    \hspace{1em}Mixed         & 5  &  51 &  3,365 & 39.9\% \\
    \hspace{1em}English       & 11 & 136 &  6,812 & 30.1\% \\
    \midrule
    \multicolumn{5}{l}{\textit{CULP}} \\
    \addlinespace[2pt]
    \hspace{1em}Lakota        & 1 &  24 &    142 & 100.0\% \\
    \hspace{1em}Mixed         & 12 & 344 & 13,552 &  26.5\% \\
    \hspace{1em}English       & 1  &  35 &    967 &   6.0\% \\
    \midrule
    \textbf{Total Corpus}     & \textbf{44} & \textbf{889} & \textbf{45,493} & \textbf{55.3\%} \\
    \bottomrule
  \end{tabular}
  \caption{Corpus overview across orthographic and language subsets. Lakota(\%) is calculated using langdetect.}
  \label{tab:data_makeup}
\end{table}

%% file: sections/ocr.tex
\section{Benchmarking OCR Approaches \& Multimodal Post-Correction}
\label{sec:ocr}

In this section, we establish the performance of zero-shot and training-based OCR systems on a gold-standard benchmark; approaches include supervised fine-tuning (SFT) directly on the recognition task or on a post-correction task where the trained model corrects a noisy first-pass. We introduce an approach for OCR post-correction which utilizes VLMs to ground the text correction in the visual prior.
We restrict our evaluation to small vision-language models with the intent that they can run locally on standard CPU hardware in seconds, eliminating the cost, cloud dependencies, and potential data privacy concerns associated with commercial APIs. We evaluate primarily open-weight models with the exception of Transkribus, which is tested in order to directly compare to the approach currently used by community partners.

\paragraph{Evaluation metrics}
For all experiments, we evaluate OCR performance using character error rate (CER) and, secondarily, word error rate (WER) \citep{Levenshtein1965BinaryCC}. 
CER is derived from the Levenshtein distance, ED:
\begin{equation*}
    \text{CER} = \frac{\text{ED}(y^*, \hat{y})}{|y^*|} = \frac{S + D + I}{N}
\end{equation*}
where $S, D, \text{ and } I$ are the number of character-level substitutions, deletions, and insertions required to transform $y^*$ into $\hat{y}$, and $N$ is the total number of characters in the reference string. 
During evaluation, we normalize whitespace and apply Unicode NFC normalization to ensure all characters are consistently precomposed before comparison.

\subsection{Zero-shot OCR} 
We first evaluate the zero-shot performance of several state-of-the-art OCR models: GLM-OCR \citep{duan2026glm}, Qwen3-VL-2B \citep{bai2025qwen3vltechnicalreport}, dots.ocr \citep{li2025dotsocrmultilingualdocumentlayout}, Nanonets OCR2 \citep{Nanonets-OCR2}, Tesseract \citep{smith2007overview}, and Transkribus. \footnote{VLM-based OCR models were selected from \url{https://www.ocrarena.ai/leaderboard}.} We run inference at the line level and calculate micro-CER across the entire dataset to weight errors proportionally. Results are shown in \cref{tab:zeroshot}. Notably, with only 0.9B parameters, GLM-OCR achieves the second best CER, only outperformed by Transkribus' Text-Titan II\footnote{\url{https://www.transkribus.org/models/text-titan-ii}}, a proprietary model. We find that the majority of errors are missing diacritics rather than misidentifications of the base character (\cref{appx:diacritic_base}).

\input{figs/zeroshot}

\subsection{Language-specific Fine-tuning }  
 We evaluate several training-based approaches to improve OCR quality over zero-shot. We focus on post-correction because a draft transcript provides scaffolding, which ideally reduces the human-annotated data required to adapt models to the task.

\paragraph{Data splits}
Due to the small dataset size, we split the 44 transcribed pages using stratified 10-fold cross-validation. Data splits are at the page level to prevent leakage across lines from the same page. Stratification balances the proportion of CULP and Deloria pages across folds as well as Lakota, English, and Mixed text. Following \citet{Rijhwani2020-aw}, each split allocates eight folds for training, one for validation, and one for testing. Each test fold contains 84--95 lines of text, accounting for 9.5\% to 10.7\% of the total dataset.

\paragraph{Multimodal post-correction}
In contrast to directly fine-tuning VLMs for OCR or using only a language model for post-correction, we provide the first-pass output as well as the image to ground the correction in both modalities. We refer to this approach as $\text{VLM}_{\text{ZS}} + \text{VLM}_{\text{POST}}$, where VLM is replaced with the specific model used in reported results.

Let $I_i$ denote the input line crop and  $\hat{y}_i^{\text{zs}}$ the initial zero-shot OCR transcription. The corrected text $\hat{y}_i$ is then: 
\[
\hat{y}_i = \arg\max_{y} P\big(y \mid I_i,\, \hat{y}_i^{\text{ZS}};\, \theta_{\text{post}}\big)
\]

We fine-tune respective VLMs using low-rank adapters \citep[LoRA; ][]{hu2021loralowrankadaptationlarge} with $r=16, \alpha=32$, learning rate $1e^{-4}$, and effective batch size of 8. For efficiency purposes, we constrain line crops to a max size of 768 pixels.

 \paragraph{Baselines} We compare multimodal post-correction against the following paradigms: 

\begin{itemize}\setlength{\itemsep}{0pt}\setlength{\parsep}{0pt}
    \item $\text{VLM}_{\text{ZS}} + \text{LSTM}_{\text{POST}}$: Sequence-to-sequence character-level post-correction \cite{Rijhwani2020-aw}.
    \item $\text{VLM}_{\text{ZS}} + \text{ByT5}_{\text{POST}}$: Byte-level neural language model post-correction \cite{lofgren2024post}.
    \item $\text{VLM}_{\text{ZS}} + \text{LLM}_{\text{POST}}$: Textual post-correction utilizing a large language model, Qwen3-1.7B \citep{yang2025qwen3technicalreport}.
    \item $\text{VLM}_{\text{SFT}}$: Direct end-to-end supervised fine-tuning on line crops \cite{chung2025fine}.
    \item $\text{Transkribus}_{\text{SFT}}$: Supervised fine-tuning of Transkribus' PyLaia base model directly on line crops.
\end{itemize}

Additional baseline implementation details can be found in \cref{appx:baseline_details}.

\input{figs/main}

\subsection{Results} 
Performance of each of the studied approaches with the three best performing zero-shot VLMs as well as Transkribus can be seen in \cref{fig:main}. Vision-grounded methods, including both multimodal post-correction (3.9--4.7\% CER) and direct fine-tuning (4.2--4.9\% CER), cut overall errors roughly fourfold compared to zero-shot baselines, significantly outperforming text-only post-correction and Transkribus (12.2\%). Corrected errors are mostly concentrated in Lakota text, where multimodal fine-tuning drops error rates from 32.6\% down to 6.2--6.7\%, compared to smaller gains on English. Across languages, multimodal post-correction and direct fine-tuning perform similarly overall ($p > 0.05$), though multimodal post-correction on GLM-OCR reaches the lowest corpus-wide error rate at 3.9\%, narrowly outperforming direct fine-tuning ($p = 0.013$). Interestingly, multimodal post-correction strongly outperforms direct SFT on mixed-language text, achieving 3.98\% vs.\ 8.82\% CER on the Deloria dataset (see \cref{appx:main_full}). This disparity may be due to hallucinations or omissions (e.g., outputting only the English), or because segmentation granularity can vary more within the mixed language texts (\cref{appx:transcription}). While these results demonstrate that language-specific finetuning of VLMs significantly improves OCR performance, they rely on idealized conditions. Next, we explore, through both additional experiments and analysis, how data- and language-specific considerations may affect performance.

%% file: figs/zeroshot.tex
\begin{table}[htbp]
  \centering
  \small
  \begin{tabular}{lrr}
    \toprule
    \textbf{Model}     & \textbf{CER (\%)} & \textbf{WER (\%)} \\
    \midrule
    GLM-OCR (0.9B)     & 17.09    & 50.33             \\
    Qwen3-VL (2B)      & 18.06             & 51.76             \\
    Dots.ocr (1.7B)      & 18.06             & \textbf{46.99}    \\
    Nanonets OCR2 (3B) & 20.32             & 67.72             \\
    Tesseract v5 (lat) & 32.58             & 68.65             \\
    Transkribus (Text Titan II) & \textbf{15.73}             & 47.65             \\
    \bottomrule
  \end{tabular}
  \vspace{0.5em}
  \caption{Overall zero-shot OCR performance on all transcribed pages.}
  \label{tab:zeroshot}
\end{table}

%% file: figs/main.tex
\begin{figure*}[h]
  \centering
  \includegraphics[width=\textwidth]{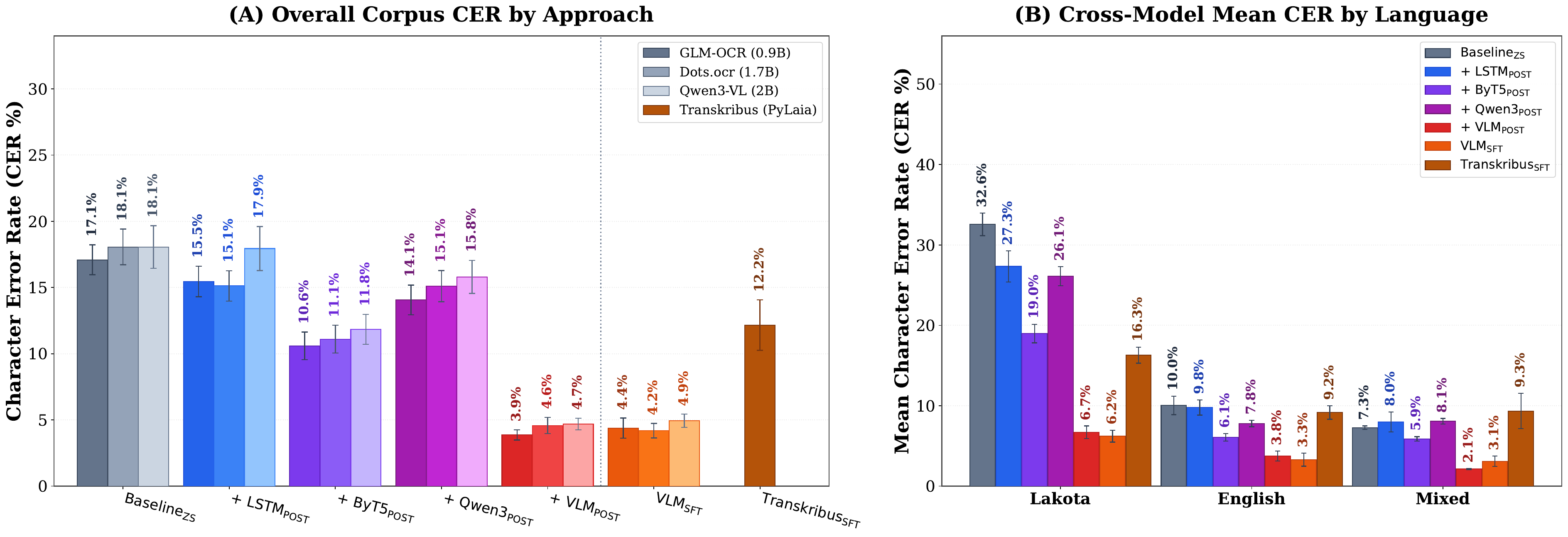}
  \caption{CER of different OCR approaches across small VLMs with direct finetuning and post-correction.}
  \label{fig:main}
\end{figure*}

%% file: sections/analysis.tex
\section{Realistic OCR for Lakota} 
Results on curated OCR benchmarks are informative, but they often ignore the practical constraints of real archival projects. In low-resource settings, these constraints include limited time for manual transcription, page segmentation challenges, mixed orthographies, and document degradation. In this section, we examine how each of these factors influences model performance and may alter design choices.

\subsection{Sample Efficiency Under Limited Transcription Budget} While OCR scaling laws have been studied from a parameter standpoint in high-resource languages \cite{rang2023empirical}, the constraint in low-resource settings tends to be human transcription time and effort. To determine the minimum manual transcription needed for meaningful adaptation, we investigate fine-tuning sample efficiency starting from an extreme low-resource baseline: a single human-transcribed page. We observe scaling behavior from one page up to our full training size of roughly 35 pages.

We evaluate each training budget $n$ across all training-based approaches using 10-fold cross-validation. To ensure reliable estimates at all budgets, we draw $k = 10$ independent random subsets of size $n$ from each training fold, yielding $100$ runs per model-budget pair. Specifically, for each of the $k$ random seeds, we fix a random permutation of the training pages and select the first $n$ elements, ensuring that smaller budgets are nested subsets of larger ones ($\mathcal{D}_n \subset \mathcal{D}_{n'}$ for $n < n'$). All reported CER values are averaged across these $100$ runs. Regardless of training budget, the models are evaluated on the same validation and test sets. 
\input{figs/scaling}

As show in \cref{fig:scaling}, text-based post-correction approaches struggle under tiny training budgets, only breaking zero-shot performance at 18 pages of training data. On the other hand, fine-tuned VLMs for OCR post-correction see performance improvements with as little as one page. By four pages, CER dips under 10\% and by nine, the model reaches 98.9\% of its performance trained on the full dataset. An edit operation error analysis demonstrates differing failure modes as well. Seq2Seq post-correction models predominantly suffer from deletion errors, while vision-grounded models exhibit mainly substitutions (\cref{appx:edit_ops}).

The amount of training data to reach adequate performance on a mixed English-Lakota corpus is surprisingly small. We hypothesize this is because VLMs can be used to classify base characters with over 97\% accuracy, so post-correction essentially becomes a diacritic restoration task that is strongly supported by vision grounding. A breakdown by language confirms that performance gains are concentrated heavily in Lakota text, whereas English and mixed lines exhibit flatter CER curves given the VLM's strong baseline English capability (see \cref{appx:scaling_by_lang})

Nonetheless, the gaps between $\text{GLM}_{\text{ZS}} + \text{GLM}_{\text{POST}}$ and $\text{GLM}_{SFT}$ are likely not large enough to justify the extra step the post-correction pipeline introduces, unless trained on only 1-2 pages of data. 
For practical Lakota use cases, direct VLM adaptation offers a more optimal balance of strong recognition and minimal human transcription effort without necessitating a multi-step inference pipeline.

\subsection{Effects of Structural Noise} 

Most work on low-resource OCR presents results based on benchmarks that assume access to gold, hand-annotated bounding boxes to isolate recognition performance from segmentation \cite{Agarwal2024-au, agarwal2025ailla}. In practice, manually segmenting every line of a document before running OCR, although a regular part of current workflows, is incredibly time consuming. Processing larger corpora may force a trade-off between annotation effort and increased error rates from automated tools. Segmentation and recognition cannot be disentangled. In this section, we measure how much off-the-shelf automatic line segmentation degrades OCR performance compared to the manual baseline. Furthermore, we evaluate whether VLMs, which are depicted as able to take in page-level inputs \citep{duan2026glm}, can bypass this segmentation step altogether.

We utilize Kraken \cite{kiessling2019kraken} to create automatic line-level bounding boxes given a page as input. We then evaluate the OCR and post-correction models trained in \cref{sec:ocr} on the line crops produced by Kraken. For the full page recognition, we feed the entire page to GLM-OCR and Transkribus to get zeroshot results, limiting image size to 2,048 pixels. For training-based approaches, we repeat 10 fold cross-validation using page-transcription pairs rather than gold line crops. 

\cref{tab:seg_noise} shows performance degradation when using automatic segmentation. On average, the use of Kraken line crops degrades CER by 2.95\%. Qualitative inspection of these line crops reveals that errors may appear due to clipped diacritics or missing handwritten notes. When using VLMs for end-to-end page-level OCR, the zero-shot performance is degraded similarly by automatic line-level bounding boxes and whole-page OCR; however, fine-tuning approaches are much more effective at the line level. This advantage is likely due to the fact that splitting pages at the line level simply results in a larger effective sample size. 

We find almost any existing performance gap between $\text{GLM}_\text{ZS} + \text{GLM}_\text{POST}$ and $\text{GLM}_\text{SFT}$ is closed when we apply automatic segmentation. Similar to the findings of the scaling experiment, $\text{GLM}_\text{SFT}$ provides the most practical solution at 7\% error rate. To ensure we maintain realistic results, we re-evaluate $\text{GLM}_\text{ZS} + \text{GLM}_\text{POST}$ with different training budgets on automatic bounding boxes (\cref{appx:scaling_seg}). The performance degradation across training budgets remains relatively stable, indicating automatic segmentation is a consistent penalty regardless of training data size. 

\input{figs/segmentation_analysis}

\subsection{Orthography-specific Models} 
Many indigenous languages do not have a single standardized writing system. Instead, they may have multiple orthographies created by linguists and community-led efforts over time. Consequently, physical archives can span evolving writing systems; historical manuscripts often preserve older orthographies (e.g., Deloria) with diacritics that differ from the more contemporary systems, such as CULP. Because documents compiled by the same linguist or archivist often share specific orthographic conventions, this raises the question of whether separate models should be trained for distinct document subsets that share an orthography, or unified into a single model.

We train both $\text{GLM}_{\text{SFT}}$ and $\text{GLM}_{\text{ZS}} + \text{GLM}_{\text{POST}}$ on the entire corpus or only one set of documents that share the same orthography. This yields an orthography-specific expert and a unified model trained on both CULP and Deloria. We then evaluate each model on their respective corpora to test if the orthography-specific experts outperform the unified model.

Interestingly, we find that training document-specific models does not always result in better OCR performance compared to training a single model across both orthographies (\cref{tab:ocr_orthography_comparison}). For the CULP documents, the models trained on the unified training corpus outperform the CULP-expert models. For the Deloria subset, the model trained on Deloria documents only outperforms the unified approach on Lakota text, but degrades the overall performance; utilizing the document specific model may only be feasible if pages can be separated by language beforehand. For Lakota orthographies, it appears that having a larger sample size, even when it mixes orthographies, results in superior performance.

\input{figs/doc_experts}

\subsection{Document Degradation}
For Lakota, many documents are historical and preserved outside of archives, resulting in scans of yellowed paper and faded ink or pencil. 
This raises a design choice: is it better to let a VLM transcribe directly, or does feeding it the initial OCR transcript provide useful scaffolding for faint characters? 

To test this, we stratify document lines into low, medium, and high Weber contrast \cite{peli1990contrast} tiers ($C_w = \frac{\Delta L}{L}$) using the Deloria collection, and report CER across model configurations in \cref{tab:contrast}. Examples of low and high contrast crops can be found in \cref{appx:contrast_ex}. We also consider $\text{ByT5}_\text{POST}$, the best performing text-based post-correction method, as a baseline unaffected by image quality outside of the zero-shot output. 

Counterintuitively, lower contrast tiers have lower zero-shot overall CER due to them also containing a higher proportion of English text; the low-contrast tier contains 40.1\% English where the medium- and high-contrast tiers only contains 27.1 and 11.5\% respectively. Thus, we also report the results only on pages manually tagged as Lakota to control for language. On Lakota text, direct vision fine-tuning ($\text{GLM}_{\text{SFT}}$) achieves the lowest error rate in high contrast (5.20\% CER) and remains robust even when contrast drops. While multimodal post-correction achieves the lowest overall error under heavy degradation at 4.75\%, direct fine-tuning proves more effective for Lakota text in the low-contrast setting. 

\input{figs/doc_contrast}

%% file: figs/scaling.tex
\begin{figure}[!b]
  \centering
  \includegraphics[width=\columnwidth]{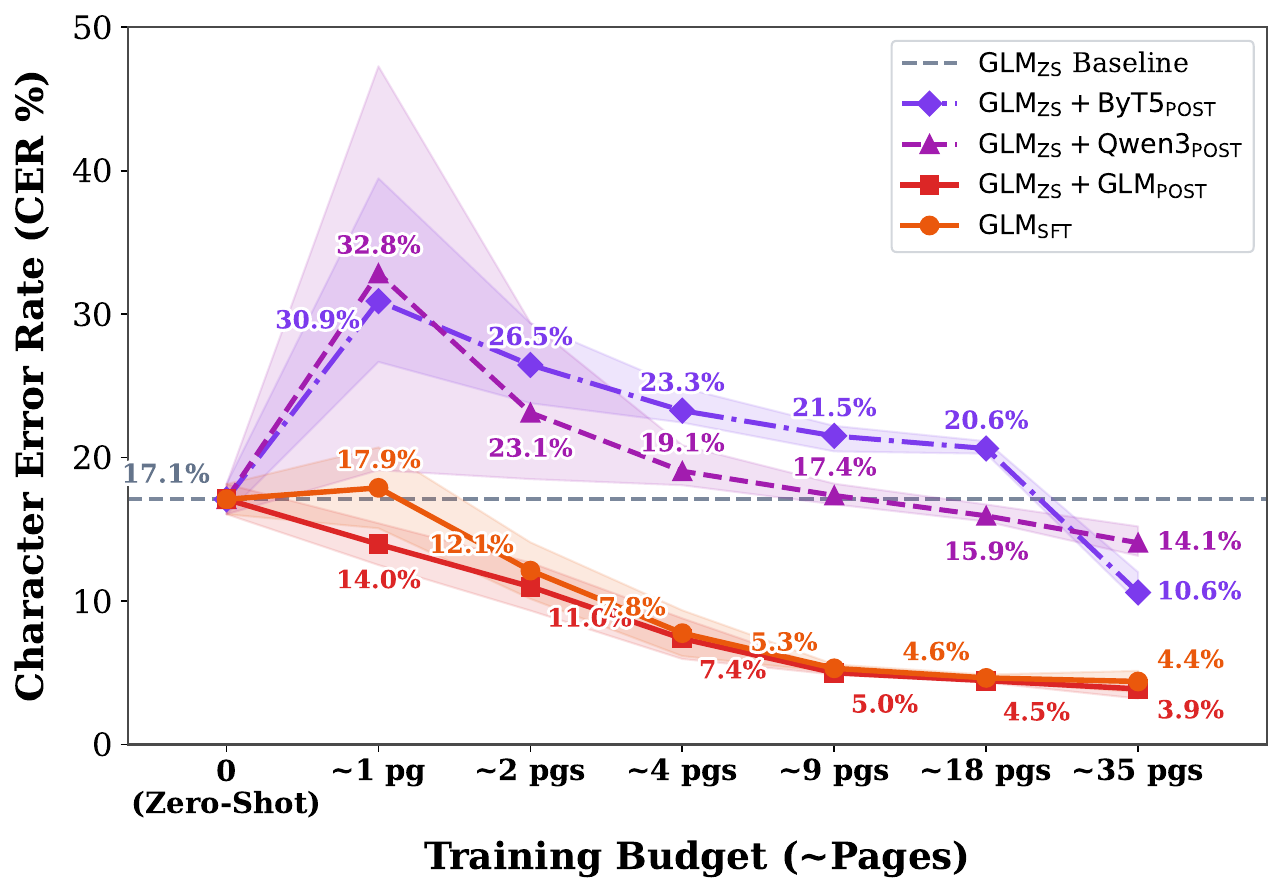}
  \caption{Solid curves show the average Character Error Rate (CER) across 10 random training-set selections under 10-fold cross-validation (100 trained models per point). Shaded regions show the 95\% confidence intervals across the different random training subsets.}
  \label{fig:scaling}
\end{figure}

%% file: figs/segmentation_analysis.tex
\begin{table}[t]
  \centering
  \small
  \renewcommand{\arraystretch}{1.15}
  \vspace{4pt}
  \begin{tabular*}{\columnwidth}{@{\extracolsep{\fill}}l rrr@{}}
    \toprule
    \textbf{Approach} & \textbf{Gold} & \textbf{Kraken} & \textbf{Full Page} \\
    \midrule
    $\text{GLM-OCR}_{\text{ZS}}$ & 17.09 & 19.82 {\scriptsize (+2.73)} & 20.06 {\scriptsize (+2.97)} \\
    $\text{Transkribus}_{\text{ZS}}$\textsuperscript{\textdagger} & 15.73 & --- & 18.20 {\scriptsize (+2.47)} \\
    \midrule
    $\text{GLM}_{\text{ZS}} + \text{LSTM}_{\text{POST}}$ & 15.46 & 19.40 {\scriptsize (+3.94)} & --- \\
    $\text{GLM}_{\text{ZS}} + \text{LLM}_{\text{POST}}$  & 14.06 & 16.76 {\scriptsize (+2.70)} & --- \\
    $\text{GLM}_{\text{ZS}} + \text{ByT5}_{\text{POST}}$ & 10.59 & 13.24 {\scriptsize (+2.65)} & --- \\
    \midrule
    $\text{GLM}_{\text{SFT}}$ & 4.39 & 7.00 {\scriptsize (+2.61)} & \textbf{11.97} {\scriptsize (+7.58)} \\
    $\text{GLM}_{\text{ZS}} + \text{GLM}_{\text{POST}}$ & \textbf{3.87} & \textbf{6.95} {\scriptsize (+3.08)} & 12.25 {\scriptsize (+8.38)} \\
    \bottomrule
  \end{tabular*}
  \caption{CER and degradation ($\Delta$) of OCR performance relative to gold bounding boxes. SFT and post-correction methods use GLM-OCR. Omitted entries (---) are not applicable to the respective pipeline or were omitted due to commercial constraints.}
  \label{tab:seg_noise}
\end{table}

%% file: figs/doc_experts.tex
\begin{table}[h]
  \centering
  \small
  \renewcommand{\arraystretch}{1.12}
  \begin{tabular*}{\columnwidth}{@{\extracolsep{\fill}}l rr@{}}
    \toprule
    \textbf{Orthography} & \textbf{Lakota} & \textbf{Corpus} \\
    \midrule
    \multicolumn{3}{l}{\textit{Deloria}} \\
    ~~$\text{GLM}_{\text{SFT}}$ (Deloria Expert) & 5.39 & 6.85 \\
    ~~$\text{GLM}_{\text{ZS}} + \text{GLM}_{\text{POST}}$ (Deloria Expert) & \textbf{5.14} & 6.25 \\
    ~~$\text{GLM}_{\text{SFT}}$ (Unified) & 5.44 & 4.27 \\
    ~~$\text{GLM}_{\text{ZS}} + \text{GLM}_{\text{POST}}$ (Unified) & 5.66 & \textbf{3.81} \\
    \midrule
    \multicolumn{3}{l}{\textit{CULP}} \\
    ~~$\text{GLM}_{\text{SFT}}$ (CULP Expert) & 3.75 & 8.42 \\
    ~~$\text{GLM}_{\text{ZS}} + \text{GLM}_{\text{POST}}$ (CULP Expert) & 3.60 & 7.65 \\
    ~~$\text{GLM}_{\text{SFT}}$ (Unified) & \textbf{3.48} & 4.27 \\
    ~~$\text{GLM}_{\text{ZS}} + \text{GLM}_{\text{POST}}$ (Unified) & 5.22 & \textbf{3.81} \\
    \bottomrule
  \end{tabular*}
  \caption{CER of models trained on orthography-specific subsets versus unified corpus training. The first column indicates the performance on Lakota text only within the specified orthography, and the Corpus column indicates the performance of all text in the orthography.}
  \label{tab:ocr_orthography_comparison}
\end{table}

%% file: figs/doc_contrast.tex
\begin{table}[t]
    \centering
    \small
    \renewcommand{\arraystretch}{1.15}
    \begin{tabular}{@{}l r r@{}}
        \toprule
        \textbf{Configuration} & \textbf{Lak} & \textbf{Overall} \\
        \midrule
        \multicolumn{3}{@{}l}{\textit{High Contrast ($C_w > 0.61$)}} \\
        ~~$\text{GLM}_{\text{ZS}}$ & 30.88 & 28.12 \\
        ~~$\text{GLM}_{\text{ZS}} + \text{ByT5}_{\text{POST}}$ & 16.87 & 15.18 \\
        ~~$\text{GLM}_{\text{SFT}}$ & \textbf{5.20} & \textbf{4.97} \\
        ~~$\text{GLM}_{\text{ZS}} + \text{GLM}_{\text{POST}}$ & 5.52 & 5.24 \\
        \midrule
        \multicolumn{3}{@{}l}{\textit{Medium Contrast ($0.58 < C_w \le 0.61$)}} \\
        ~~$\text{GLM}_{\text{ZS}}$ & 30.97 & 24.64 \\
        ~~$\text{GLM}_{\text{ZS}} + \text{ByT5}_{\text{POST}}$ & 20.49 & 16.36 \\
        ~~$\text{GLM}_{\text{SFT}}$ & 5.88 & \textbf{5.04} \\
        ~~$\text{GLM}_{\text{ZS}} + \text{GLM}_{\text{POST}}$ & \textbf{5.82} & 5.12 \\
        \midrule
        \multicolumn{3}{@{}l}{\textit{Low Contrast ($C_w \le 0.58$)}} \\
        ~~$\text{GLM}_{\text{ZS}}$ & 31.85 & 20.51 \\
        ~~$\text{GLM}_{\text{ZS}} + \text{ByT5}_{\text{POST}}$ & 16.74 & 11.92 \\
        ~~$\text{GLM}_{\text{SFT}}$ & \textbf{5.54} & 7.03 \\
        ~~$\text{GLM}_{\text{ZS}} + \text{GLM}_{\text{POST}}$ & 6.11 & \textbf{4.75} \\
        \bottomrule
    \end{tabular}
    \caption{CER (\%) across Weber optical contrast tiers on historical Deloria manuscripts.}
    \label{tab:contrast}
\end{table}

%% file: sections/conclusion.tex
 \section{Conclusion}
Our findings show that VLMs are more effective than text-based post-correction for transcribing Lakota. Fine-tuned VLMs scale quickly with small training budgets, and can generalize across orthographies. Because these models run locally on CPU, community members can transcribe documents without relying on proprietary tools. For ongoing transcription, direct fine-tuning provides the most practical solution, though multimodal post-correction may be more robust to mixed-language texts or varied segmentation granularity. However, structural noise remains a hurdle: whole-page VLM inputs perform poorly, and automatic segmentation tools introduce errors like clipped diacritics. Pairing automatic segmentation with a fine-tuned VLM gives a realistic result of 7.00\% CER, an 59\% increase from the result on gold bounding boxes. By bringing attention to current shortcomings and making design choices around language-specific considerations, we hope to remove as much friction in the transcription process as possible. If technology is going to support language revitalization by making archival materials accessible for community members, it must be built for these real-world conditions, ensuring that technical evaluations actually serve ongoing reclamation efforts. 

%% file: sections/ethics.tex
\section*{Ethical Considerations}
\label{sec:ethics}

While the materials used in this study exist within the public domain, specific sources will not be linked or publicly released. This decision is to respect the perspective of tribal members who maintain that such materials should not be available to the general population. Furthermore, the broader archival initiative this work intends to support is intended for internal tribal access only. Research on Indigenous languages must always center community governance. For many North American tribes, including the Lakota, restricting access is a response to historical and ongoing colonial exploitation.

Language technologies can risk cementing external standards over community autonomy. The CULP orthography is a pre-cursor to the "Standard" Lakota Orthography (SLO). Created by the Lakota Language Consortium (LLC), the SLO is controversial for a few reasons: (a) many community members object to defining a "standard" for the diverse spoken language, and (b) the LLC's status as an outside entity that sought copyrights over cultural materials. SLO was adopted by communities such as Standing Rock for its systematic approach being easier to use in educational settings, who reclaim it by referring to it as the "suggested" orthography.

%% file: sections/limitations.tex
\section*{Limitations}
While the goal of our work is to support the revitalization and digitization effort of the Lakota community, this also means the findings may not be generalizable to other languages and communities. 
Lakota is a Latin-script language. Despite uncommon diacritics, this work does not represent outcomes for languages in non-Latin scripts, or scripts not supported by standard character encodings like Unicode.
Furthermore, we assess recognition performance using CER and WER, but a reduction in these metrics does not automatically translate to reduced proofreading effort, so verifying true reductions in human labor will require human-computer interaction studies. 
In our realistic pipeline experiments, models were evaluated on automated line segmentations from Kraken while trained solely on gold bounding boxes; fine-tuning directly on transcriptions aligned to automated bounding boxes may better condition models to segmentation noise. Additionally, our dataset is restricted to two orthographies and mostly cleanly typeset materials, leaving purely handwritten manuscripts and more than ten Lakota orthographies unexamined. Our work also does not differentiate across mutually intelligible dialects like Dakota. Finally, for consistency, we train the Transkribus PyLaia model using the same gold bounding boxes as the other models. However, Transkribus offers its own segmentation system that may offer better results when natively paired with the recognition model. We do not explore this due to the additional effort of re-annotating the segmentations as well as cost. 

%% file: sections/acknowledgements.tex
\section*{Acknowledgments}
We would like to thank Elliot Bannister and Nacole Walker of the Iyápi Program at Standing Rock Reservation for insightful discussions on the current transcription process and use of OCR in digitization pipelines.

This material is based upon work supported by the National Science Foundation Graduate Research Fellowship Program under Grant No. DGE 2146752. Any opinions, findings, and conclusions or recommendations expressed in this material are those of the author(s) and do not necessarily reflect the views of the National Science Foundation.

%% file: sections/appendix/lakota.tex
\clearpage
\section{Additional Background on Lakota and Dakota}
\label{sec:appendix_lakota}
The Lakota language, spoken in the Northern Plains of current-day U.S. and Canada and has a few mutually intelligible languages, Lakota, Dakota, and Nakoda, all three of which are sometimes referred to as “Sioux” \citep{esteban20179}. They are commonly abbreviated as NLD among other variants. It is a polysynthetic Siouan language with high verbal morphology. The Lakota language is critically endangered, with fewer than 3,000 native speakers remaining, the majority of whom are elderly \citep{antoine2024lakota}. Numerous revitalization efforts, such as immersion schools, community classes, language camps, etc., have been launched across regions where it is traditionally spoken to encourage language use \citep{Walsh_Jungemann_Koester_2026}. 

Bare verbs, lacking affixation, convey third person singular and present tense. Any further markers for person, plural, animacy, etc. are inflected from the base verb through sets of pronominal affixation reflecting either nominative-accusative or ergative-absolutive alignment. Lakota has five oral vowels, a, e, i, o, \& u, and three nasal vowels that are ã,  ĩ,  \&  ũ  \citep{paschen2022trigger}. Lakota has twenty-seven consonants. These consonants include plain, aspirated, ejective plosives, and glottalized fricatives \citep{lakota2008new}.

\subsection{Comparison of Orthographies}
\label{sec:orthography_comparison}
To illustrate the orthographic divergence across document collections, an author who is a Lakota speaker transliterated an example sentence from the Deloria corpus into the Suggested Lakota Orthography (SLO), with which they are more familiar than the closely related CULP system (Table~\ref{tab:orthography_comparison}).

\begin{table}[ht]
\centering
\small
\begin{tabular}{@{}lp{5.2cm}@{}}
\toprule
\textbf{Orthography} & \textbf{Example} \\
\midrule
\textbf{Deloria} & Waná ak‘é t’e-blézeṡni ḣ’ąyé;’ p‘á yazą́ keyé.’ \\
\addlinespace
\textbf{SLO}     & Waŋná akhé t'é blézešni ȟ'aŋyáŋ ``pȟá yazáŋ,'' kéye. \\
\bottomrule
\end{tabular}
\caption{Parallel sentence illustrating orthographic differences between historical Deloria manuscripts and the Suggested Lakota Orthography (SLO).}
\label{tab:orthography_comparison}
\end{table}

%% file: sections/appendix/examples.tex
\section{Examples of Data and OCR Outputs}
In this section, we provide examples of both inputs and outputs to OCR to demonstrate differences between orthographies or contrast levels.

\subsection{Data Examples}
\label{appx:examples}
\begin{figure}[h]
  \centering
  \includegraphics[width=\columnwidth]{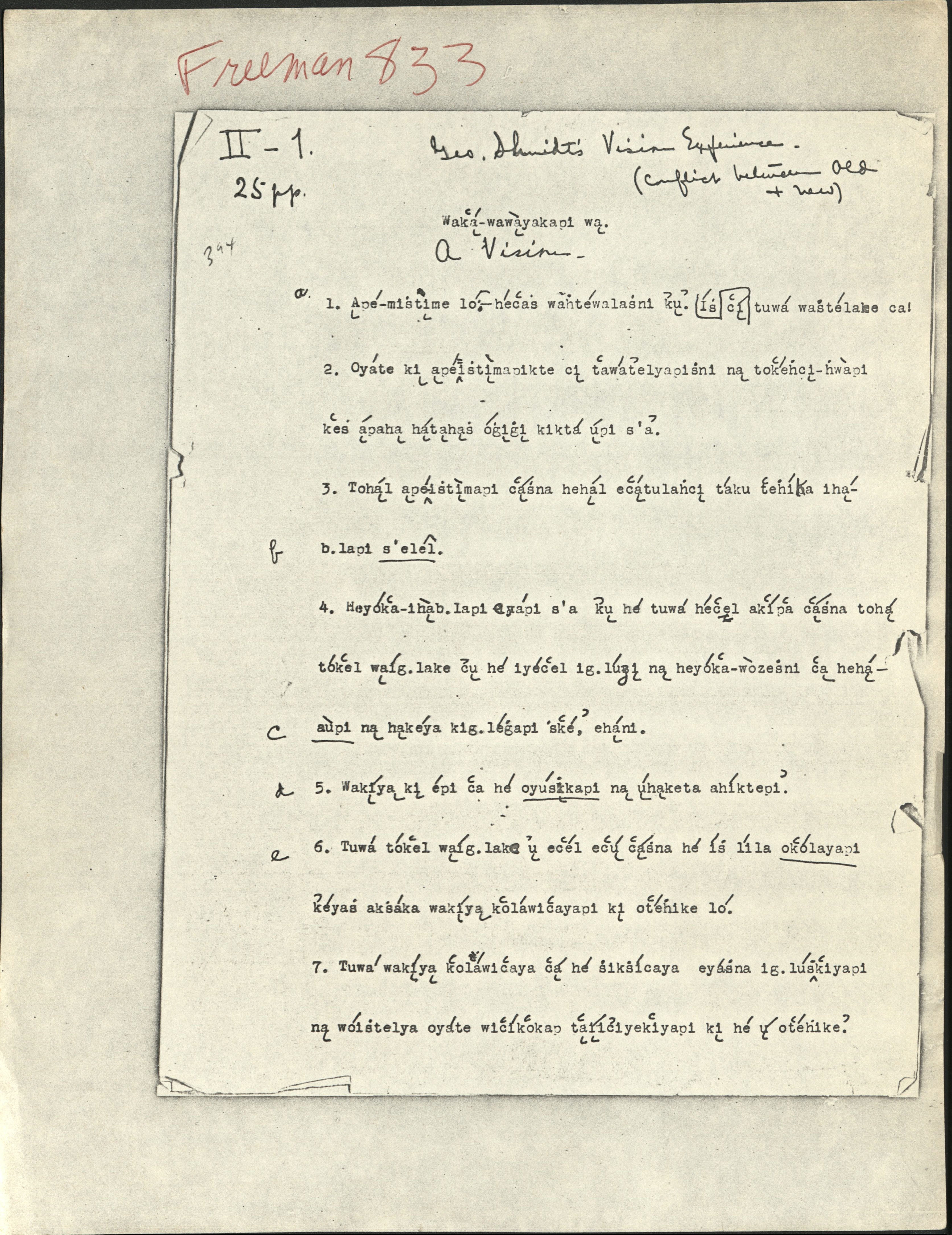}
  \caption{An example of a Lakota page from the Deloria corpus. This page highlights the handwritten diacritics as well as document degradation.}
  \label{fig:ex1}
\end{figure}
\begin{figure}[h]
  \centering
  \includegraphics[width=\columnwidth]{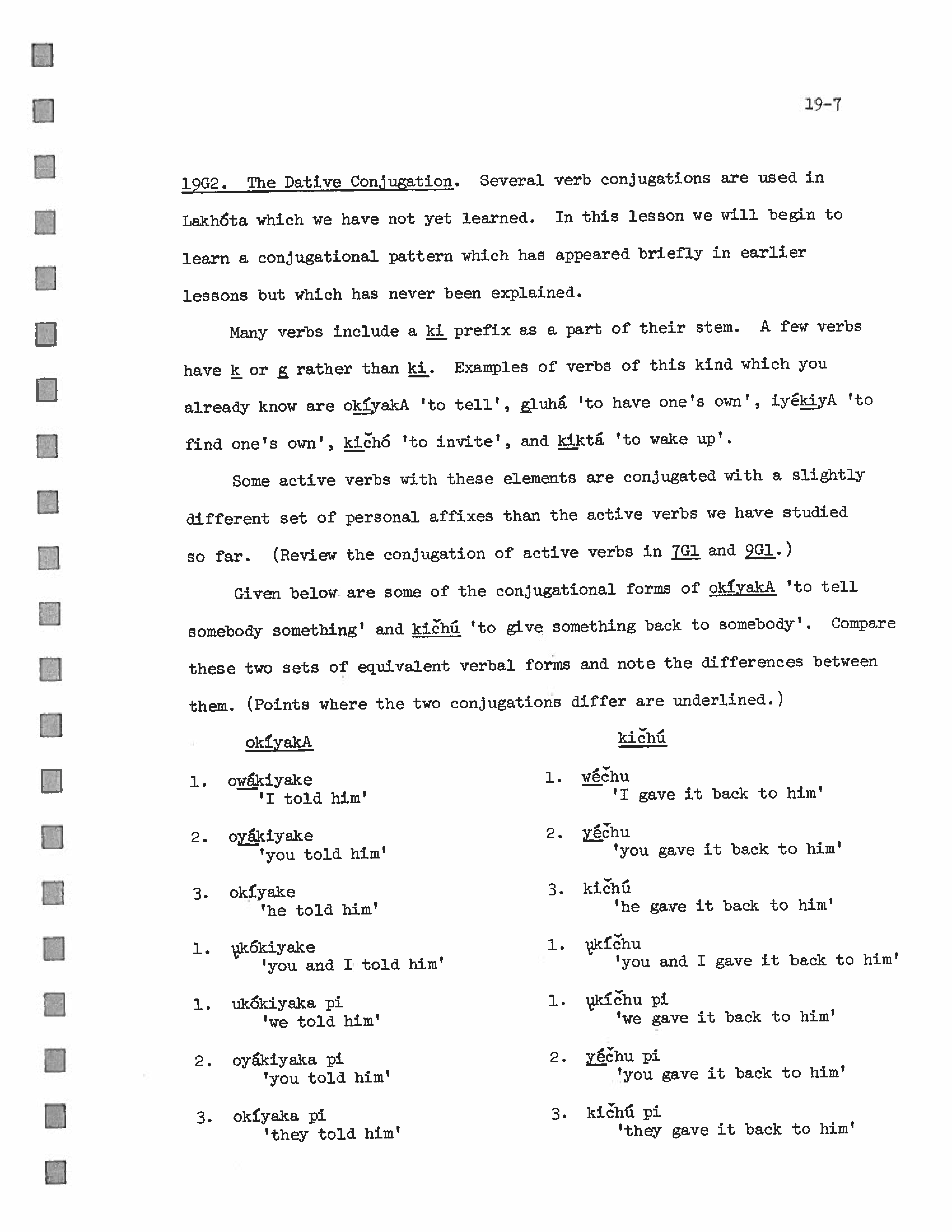}
  \caption{An example of a bilingual page from the CULP learning materials. This page highlights the clean document quality in comparison to the Deloria materials as well as the more mixed language nature of the text. Where Deloria is easily separable at the page level by language, the Lakota and English are interleaved in the learning materials. Furthermore, the document highlights structural challenges like double columns and hole punches that can confuse automatic segmentation models.}
  \label{fig:ex2}
\end{figure}
\begin{figure}[h]
  \centering
  \includegraphics[width=\columnwidth]{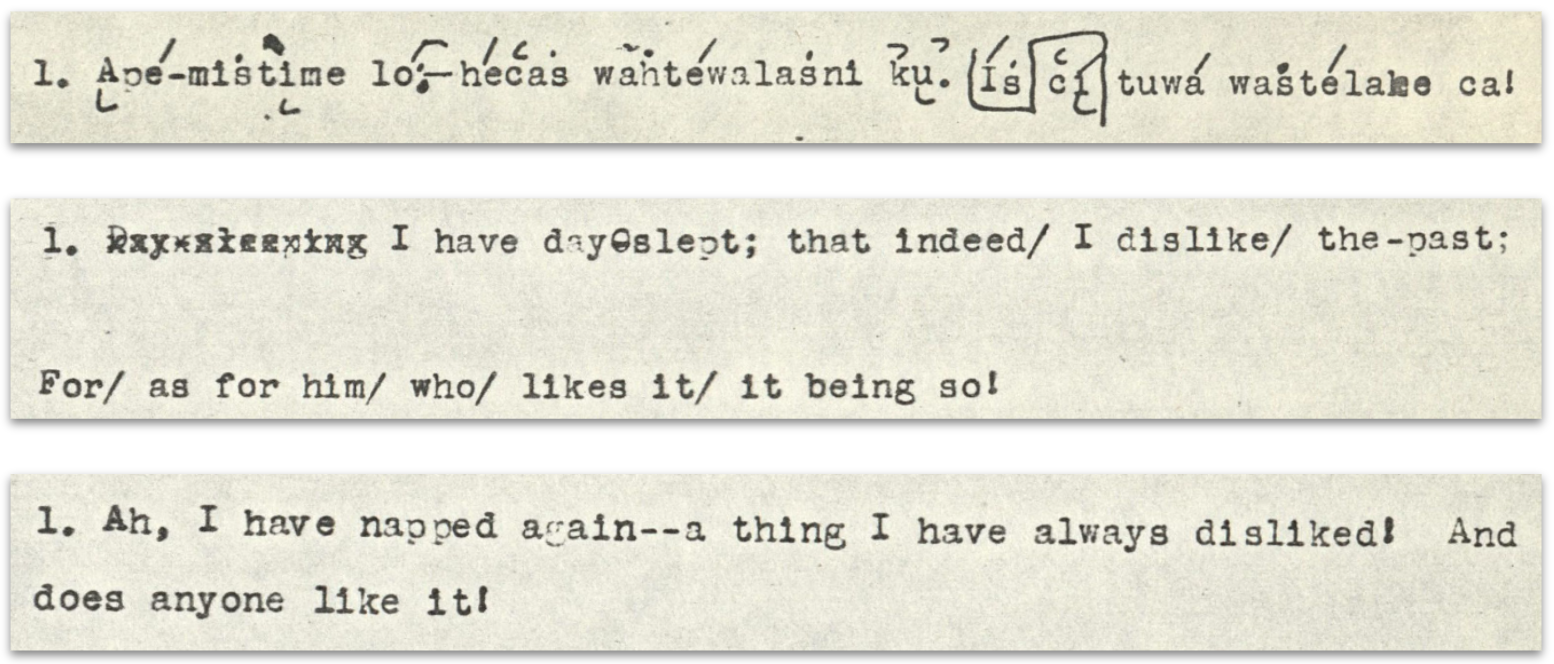}
  \caption{An example of a Lakota sentence, its word-for-word translation, and free translation in the Deloria corpus.}
  \label{fig:ex3}
\end{figure}

\subsection{Example of Low and High Contrast}
\label{appx:contrast_ex}
\begin{figure}[t]
  \centering
  \includegraphics[width=\columnwidth]{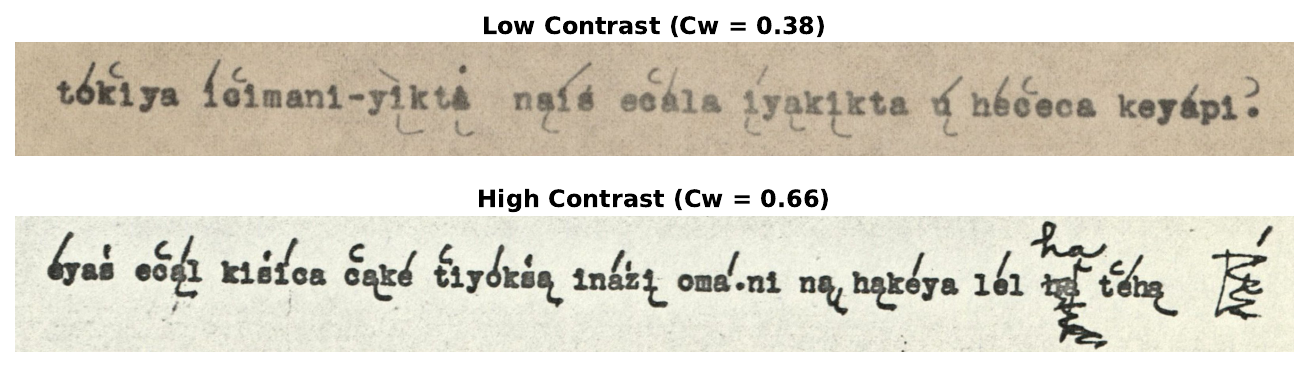}
  \caption{An example of a low and high contrast line from the Deloria documents.}
  \label{fig:contrast_ex}
\end{figure}

In \cref{fig:contrast_ex}, we show the difference between low- and high-contrast line crops. Both are taken from the Deloria corpus, but the low-contrast example exhibits significant ink fading.

\subsection{Examples of OCR Outputs}
\label{appx:outputs}
\begin{figure*}[p]
  \centering
  \includegraphics[width=\textwidth,height=0.84\textheight,keepaspectratio]{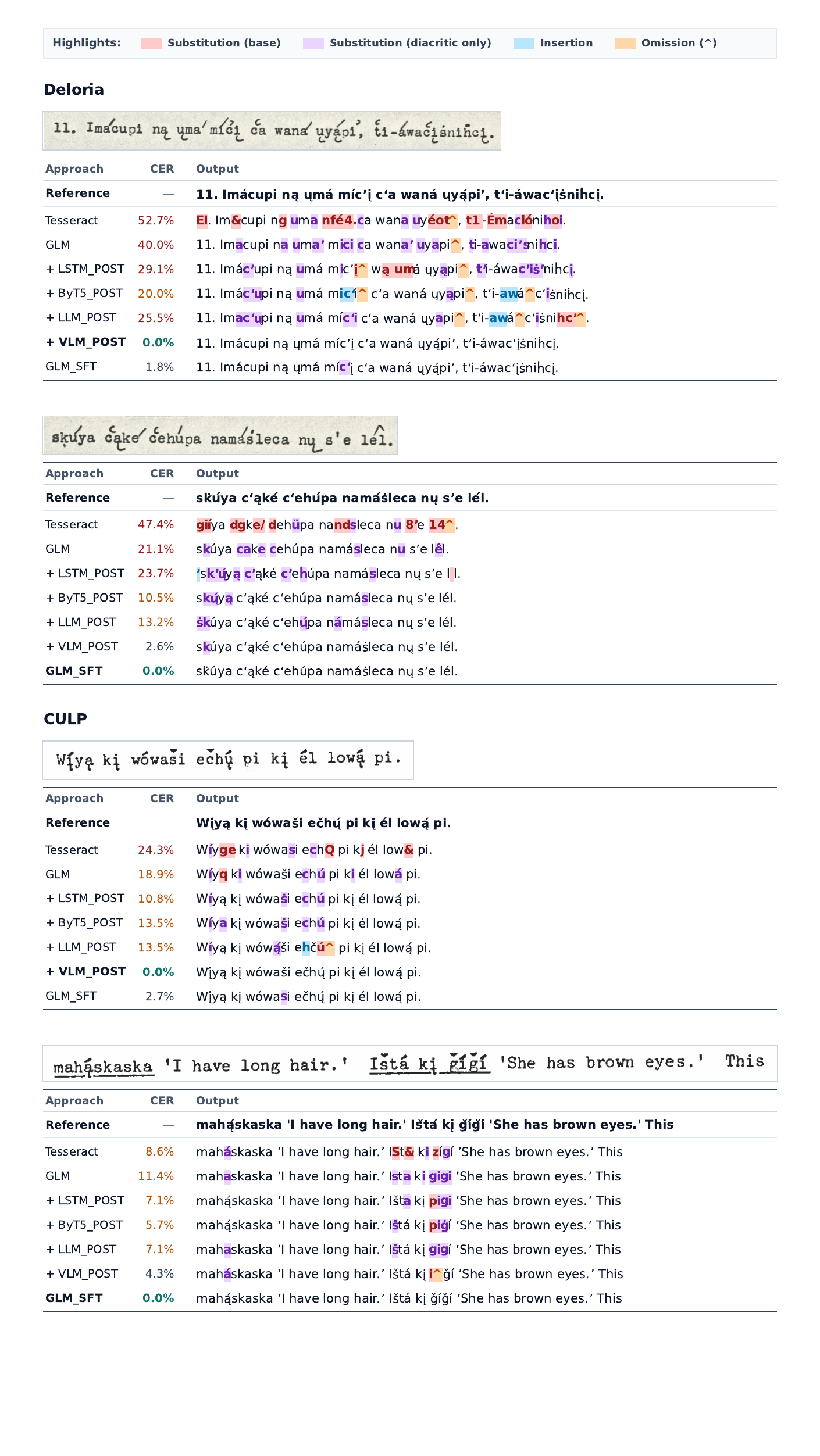}
  \caption{Examples of line crops and the corresponding transcriptions from various OCR and post-correction paradigms with GLM-OCR. We additionally include Tesseract to illustrate higher baseline error rates.}
  \label{fig:ocr_outputs}
\end{figure*}

Example outputs across various OCR baselines and post-correction paradigms are detailed in \cref{fig:ocr_outputs}.

% Full-page two-column figure on its own dedicated clean page

%% file: sections/appendix/implementation_details.tex
\section{Extended Methodology and Details}
\subsection{Baseline Implementations}
\label{appx:baseline_details}
\begin{itemize}
    \item $ \text{LSTM}_{\text{POST}}$: Implemented in DyNet using a 2-layer character-level BiLSTM encoder with attention, overage vectors, and a pointer-generator copy mechanism (diagonal loss). It undergoes 10 epochs of unsupervised character-denoised pretraining followed by supervised fine-tuning with DyNet autobatching until early stopping on validation CER. We implemented this directly using the repository provided in \citet{Rijhwani2020-aw}.
    \item $\text{ByT5}_{\text{POST}}$: Implemented using google/byt5-small/base operating         
    tokenizer-free directly on raw UTF-8 bytes prefixed with "Correct OCR: ". The model is fine-tuned with lr=$5e-4$ and effective batch size 16 under a cosine decay schedule with 10\% warmup, and early stopping on validation CER.
    \item $\text{LLM}_{\text{POST}}$: We train the LLM with identical training setup to the VLM post-correction. Implemented on Qwen/Qwen3-1.7B using PEFT LoRA ($r = 16, \alpha = 32$). It is optimized with AdamW (lr=$1e-4$, effective batch size 8, cosine schedule) and evaluated every 25 steps for early stopping on validation CER. 
    \item $\text{VLM}_{\text{SFT}}$: Implemented on zai-org/GLM-OCR with identical LoRA hyperparameters and optimization settings to our main approach, but trained purely end-to-end from image to text with zero draft text input.   
    \item $\text{Transkribus}_{\text{SFT}}$: Direct supervised fine-tuning on line crops from Transkribus' base PyLaia model, a credit based service for OCR. We train the models on Transkribus' web app training from their "base model" for 250 epochs with early stopping patience 20, and execute inference via their API. We utilize our own bounding boxes and evaluate CER identically to other models as opposed to using the reported CER from their webapp.
\end{itemize}

\clearpage
\subsection{Post-correction Augmentations} 
\label{post_corr_augmentations}
During development, we tried several augmentations to the multimodal post-correction training. In \cref{tab:draft_ensemble}, we show results when training a VLM for post-correction on its own zero-shot drafts, the best zero-shot draft, and ensembling all zero-shot drafts to increase data amount and diversity. 
\input{figs/appendix/ensemble}
We also test if increasing the diversity of the images with AutoAugment \cite{cubuk2018autoaugment}, a tool meant to increase the accuracy of vision classifiers; however, we find that using these augmentation has little to no change. In fact, it degrades accuracy from 3.87 to 3.94\%. 

We also try applying classifier-free guidance (CFG), a tool used in image generation and editing to allow for models to be more robust by randomly dropping either vision or text inputs \cite{ho2022classifier}. Without CFG, post-correction performs at 3.98\% CER on the entire corpus, and with it, it drops to 3.87\%. While the difference is still relatively small, we choose to employ CFG in our main results. This way, the post-correction system can also operate as an OCR system by dropping the noisy text input. 

\subsection{Transcription Details}
\label{appx:transcription}
Two authors annotated bounding boxes and transcribed the 44 pages in our dataset using LabelStudio (\url{https://labelstud.io/}), which has an OCR annotation preset. We did not use bounding boxes with rotation. Bounding boxes were most commonly at the line-level, but could also be smaller of larger granularity. For example, for two-column setups in the learning materials, we annotate the two sides as separate boxes rather than one line. For some of the notes in the Deloria documents, we annotate each note as a bounding box, even if it is longer than one line due to the lines being very close together. To avoid excessive copy-pasting and ensure the same diacritics across annotators, we utilized the Lakota-All-In-One keyboard from \url{https://www.languagegeek.com/siouan/keyboards/si_kbds.html}. 

%% file: figs/appendix/ensemble.tex
\begin{table}[H]
\centering
\small
\setlength{\tabcolsep}{6pt}
\begin{tabular}{lcrrrr}
\toprule
\textbf{Model} & \textbf{ZS} & \textbf{Own} & \textbf{Best} & \textbf{Ensemble} \\
\midrule
GLM-OCR(0.9B) & 17.09 & \textbf{3.87} & 3.87 & 4.05 \\
Qwen3-VL(2.0B) & 18.06 & 4.69          & 5.20 & \textbf{4.52} \\
\bottomrule
\end{tabular}
\caption{CER across draft conditioning settings. Own refers to post-correcting the zero-shot draft from the same model. Best refers to correcting the zero-shot draft of the VLM with the lowest zero-shot CER (GLM-OCR). Finally, ensemble refers to training the model on each zero-shot draft and corrected version as a separate data point. While we hypothesized that increasing the data size and diversity of errors may improve performance, the ensemble method actually makes performance worse for GLM and slightly improves Qwen3-VL. Because we use GLM-OCR for the majority of experiments, we omit ensembling. This also provides a simpler training setup.}
\label{tab:draft_ensemble}
\end{table}

%% file: sections/appendix/full_results.tex
\section{Full OCR Results}
In this section, we provide supplementary results to those presented in \cref{sec:ocr}.
\label{sec:appendix_fullresults}

\subsection{Zero-shot OCR results by language and orthography}
\label{appx:zeroshot_full}
The results in \cref{tab:zeroshot_full} show that Transkribus' Text Titan II model is consistently strongest on Lakota text across orthographies, but GLM-OCR excels for mixed text in the Deloria orthography and dots.OCR performs well on the clean mixed and English text in the learning materials.
\input{figs/appendix/zeroshot_full}

\clearpage
\subsection{Diacritic and base-character only CER}
We report base character and diacritic specific error rates (\cref{tab:diacritic_base}). As expected, the base character error rate is much lower. For example, GLM-OCR attains 2.75\% base-character accuracy, but misses 85\% of diacritics. Text-Titan II performs more strongly, likely due to pretraining on predominantly Western European languages which include diacritics like accents. 
\label{appx:diacritic_base}
\input{figs/appendix/diacritic_base}

\subsection{Full OCR results for training-based methods}
We report the full results across languages and orthographies for training-based paradigms for OCR. 
\label{appx:main_full}
\input{figs/appendix/main_full}

%% file: figs/appendix/zeroshot_full.tex
\begin{table}[H]
\centering
\scriptsize
\setlength{\tabcolsep}{2.5pt}
\textbf{(a) Deloria}\\[0.3em]
\begin{tabular}{lrrrrrr}
\toprule
 & \multicolumn{2}{c}{\textbf{Lakota}} & \multicolumn{2}{c}{\textbf{Mixed}} & \multicolumn{2}{c}{\textbf{English}} \\
\cmidrule(lr){2-3} \cmidrule(lr){4-5} \cmidrule(lr){6-7}
\textbf{Model} & \textbf{CER} & \textbf{WER} & \textbf{CER} & \textbf{WER} & \textbf{CER} & \textbf{WER} \\
\midrule

Dots OCR           & 32.38          & 88.30          & 20.89          & 44.99          & 12.02          & 28.86 \\
GLM-OCR            & 31.48          & 91.35          & \textbf{13.11} & \textbf{38.14} &  9.46          & \textbf{28.15} \\
Qwen-2B            & 33.10          & 96.36          & 14.99          & 40.77          & 10.99          & 30.55 \\
Nanonets OCR2 (3B) & 30.20          & 85.65          & 33.48          & 76.27          & 13.47          & 68.47 \\
Tesseract v5 (Lat) & 53.75          & 110.75         & 57.10          & 89.10          & 22.89          & 53.73 \\
Text Titan II        & \textbf{26.24} & \textbf{83.14} & 30.94          & 55.54          &  \textbf{9.23} & 34.01 \\
\bottomrule
\end{tabular}

\vspace{0.9em}
\textbf{(b) CULP}\\[0.3em]
\begin{tabular}{lrrrrrr}
\toprule
 & \multicolumn{2}{c}{\textbf{Lakota}} & \multicolumn{2}{c}{\textbf{Mixed}} & \multicolumn{2}{c}{\textbf{English}} \\
\cmidrule(lr){2-3} \cmidrule(lr){4-5} \cmidrule(lr){6-7}
\textbf{Model} & \textbf{CER} & \textbf{WER} & \textbf{CER} & \textbf{WER} & \textbf{CER} & \textbf{WER} \\
\midrule

Dots OCR           & 15.79          & 45.83          &  3.84          & \textbf{11.19} & \textbf{1.56} &  \textbf{5.56} \\
GLM-OCR            & 22.81          & 70.83          &  5.03          & 19.15          & 1.66          &  8.33 \\
Qwen-2B            & 20.18          & 58.33          &  4.80          & 16.01          & 2.08          &  6.67 \\
Nanonets OCR2 (3B) & 63.16          & 83.33          &  8.11          & 46.04          & 17.59         & 43.33 \\
Tesseract v5 (Lat) & 22.81          & 70.83          &  7.54          & 25.90          & 3.02          & 13.33 \\
Text Titan II      &  \textbf{8.77} & \textbf{37.50} &  \textbf{3.31} & 13.90          & 2.50          &  9.44 \\
\bottomrule
\end{tabular}
\caption{CER and WER (\%) across languages and orthographies. Plot (a) shows results for the Deloria subset of the dataset and (b) shows it for the SLO learning materials.}
\label{tab:zeroshot_full}
\end{table}

%% file: figs/appendix/diacritic_base.tex
\begin{table}[H]
\centering
\small
\begin{tabular}{lrrr}
\toprule
\textbf{Model} & \textbf{ZS} & \textbf{Diacritic}  & \textbf{Base}  \\
\midrule
Transkribus (Text Titan II) & \textbf{15.73} & \textbf{61.20} & 5.79 \\
GLM-OCR (0.9B)       & 17.09          & 85.28          & \textbf{2.75} \\
Dots.ocr (1.7B)      & 18.06          & 74.32          & 5.28 \\
Qwen3-VL (2B)        & 18.06          & 85.06          & 4.47 \\
Nanonets OCR2 (3B)   & 20.32          & 68.37          & 9.42 \\
Tesseract v5 (Lat)   & 32.58          & 84.93          & 28.14 \\
\bottomrule
\end{tabular}
\caption{Zero-shot character error rates (CER) across OCR and vision-language models, broken down by overall CER, diacritic-only CER, and base character CER (\%). }
\label{tab:diacritic_base}
\end{table}

%% file: figs/appendix/main_full.tex
\begin{table}[H]
\centering
\small
\setlength{\tabcolsep}{5.5pt}

\textbf{(a) Deloria}\\[0.3em]
\begin{tabular}{lrrrr}
\toprule
\textbf{Paradigm} & \textbf{Lak} & \textbf{Mix} & \textbf{En} & \textbf{All CER} \\
\midrule
$\mathrm{Baseline_{ZS}}$     & 32.64          & 16.67          & 11.28          & 25.65 \\
$\mathrm{+\ LSTM_{POST}}$     & 28.48          & 19.50          & 11.50          & 23.36 \\
$\mathrm{+\ ByT5_{POST}}$     & 19.21          & 12.34          &  6.81          & 15.43 \\
$\mathrm{+\ Qwen3_{POST}}$    & 24.85          & 11.85          &  8.65          & 19.82 \\
$\mathrm{+\ VLM_{POST}}$     &  6.72          &  \textbf{3.98} &  4.15          &  \textbf{5.78} \\
$\mathrm{VLM_{SFT}}$          &  \textbf{6.21} &  8.82          &  \textbf{3.64} &  5.90 \\
$\mathrm{Transkribus_{SFT}}$  & 15.48          & 16.05          &  9.15          & 13.56 \\
\bottomrule
\end{tabular}

\vspace{0.9em}
\textbf{(b) CULP}\\[0.3em]
\begin{tabular}{lrrrrr}
\toprule
\textbf{Paradigm} & \textbf{Lak} & \textbf{Mix} & \textbf{En} & \textbf{All CER} \\
\midrule
$\mathrm{Baseline_{ZS}}$     & 20.58          & 4.99           & 1.77           & 4.89           \\
$\mathrm{+\ LSTM_{POST}}$     & 21.74          & 5.45           & 1.87           & 5.33         \\
$\mathrm{+\ ByT5_{POST}}$     & 20.87          & 4.34           & 1.92           & 4.31           \\
$\mathrm{+\ Qwen3_{POST}}$    & 22.40          & 3.90           & 1.45           & 3.85           \\
$\mathrm{+\ VLM_{POST}}$     &  \textbf{4.93} &  \textbf{1.70} & 1.18           &  \textbf{1.69}  \\
$\mathrm{VLM_{SFT}}$          &  6.38          &  1.71          & \textbf{1.04}  &  1.70          \\
$\mathrm{Transkribus_{SFT}}$  &  6.96          &  4.50          & 2.20           &  4.82          \\
\bottomrule
\end{tabular}
\caption{CER across modeling paradigms for Deloria and SLO.}
\label{tab:main_full}
\end{table}

%% file: sections/appendix/additional_analyses.tex
\section{Additional Scaling Analyses}

\subsection{Scaling results by language}
\label{appx:scaling_by_lang}
\input{figs/appendix/scaling_by_lang}

\subsection{Scaling results by segmentation method}
\label{appx:scaling_seg}
\input{figs/appendix/scaling_gold_kraken}

\subsection{Edit-distance operations analysis} We analyze the scaling behavior of each OCR or post-correction paradigm using edit distance operations: insertions, substitutions, and deletions in \cref{fig:edit_ops}.
\label{appx:edit_ops}
\include{figs/appendix/edit_ops}

%% file: figs/appendix/scaling_by_lang.tex
\begin{figure}[H]
  \centering
  \includegraphics[width=\linewidth]{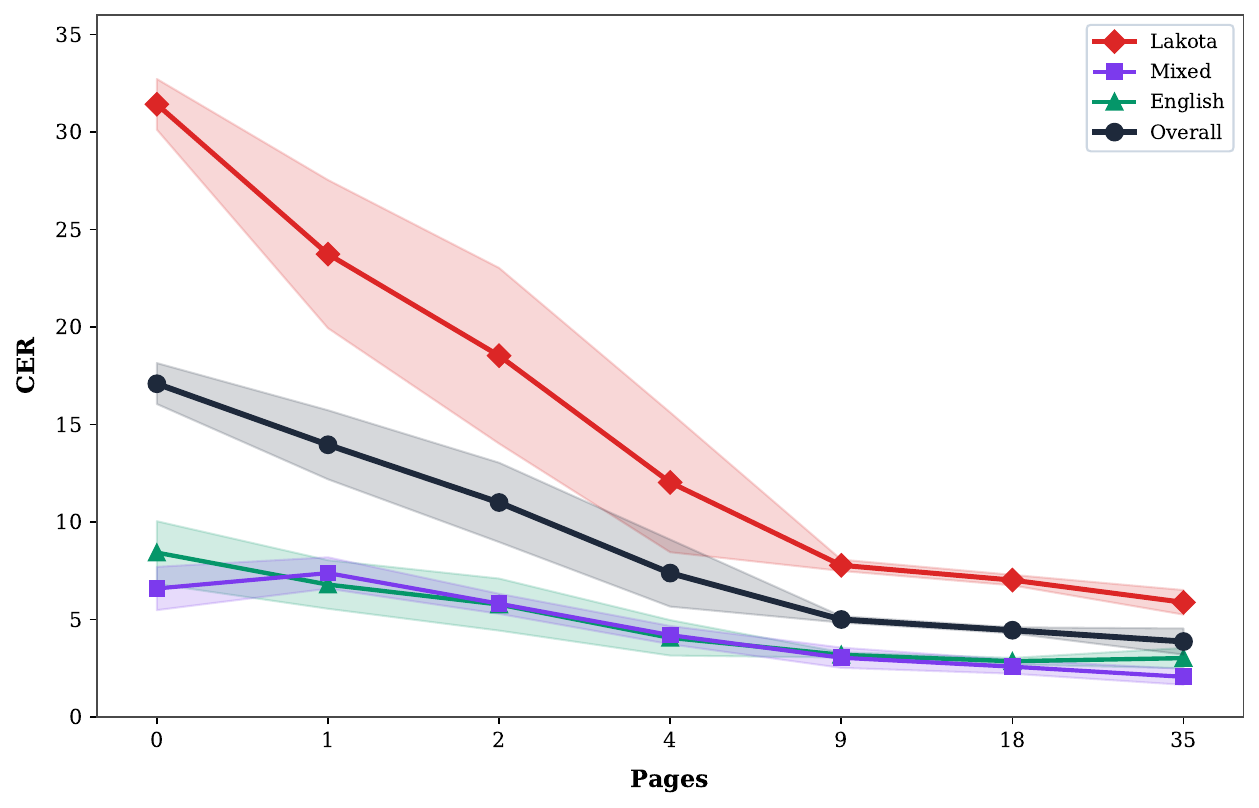}
  \caption{Results of $\text{VLM}_{\text{ZS}} + \text{VLM}_{\text{corr}}$ separated by language. The majority of gains is concentrated in Lakota whereas English and mixed language CERs improve little in comparison.}
  \label{fig:edit_ops}
\end{figure}

%% file: figs/appendix/scaling_gold_kraken.tex
\begin{figure}[h]
  \centering
  \includegraphics[width=\linewidth]{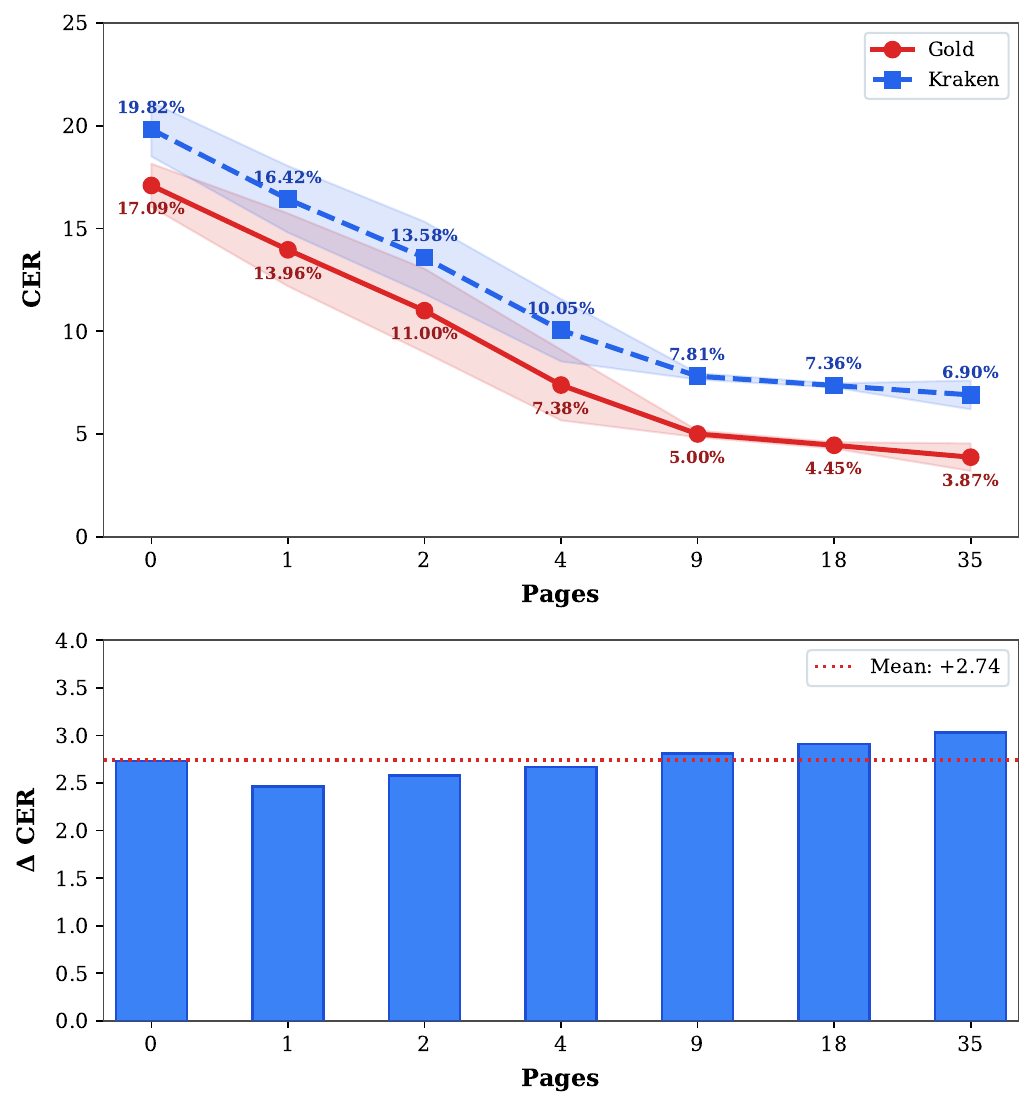}
  \caption{Results of $\text{VLM}_{\text{ZS}} + \text{VLM}_{\text{corr}}$ evaluated on gold line-crops versus line-crops produced by Kraken, the automatic segmentation model. Error deltas remain relatively constant across training budgets, indicating that training on gold bounding boxes does not provide any improvements on automatic ones.}
  \label{fig:kraken_scaling}
\end{figure}

%% file: figs/appendix/edit_ops.tex
\begin{figure*}[h]
  \centering
  \includegraphics[width=\textwidth]{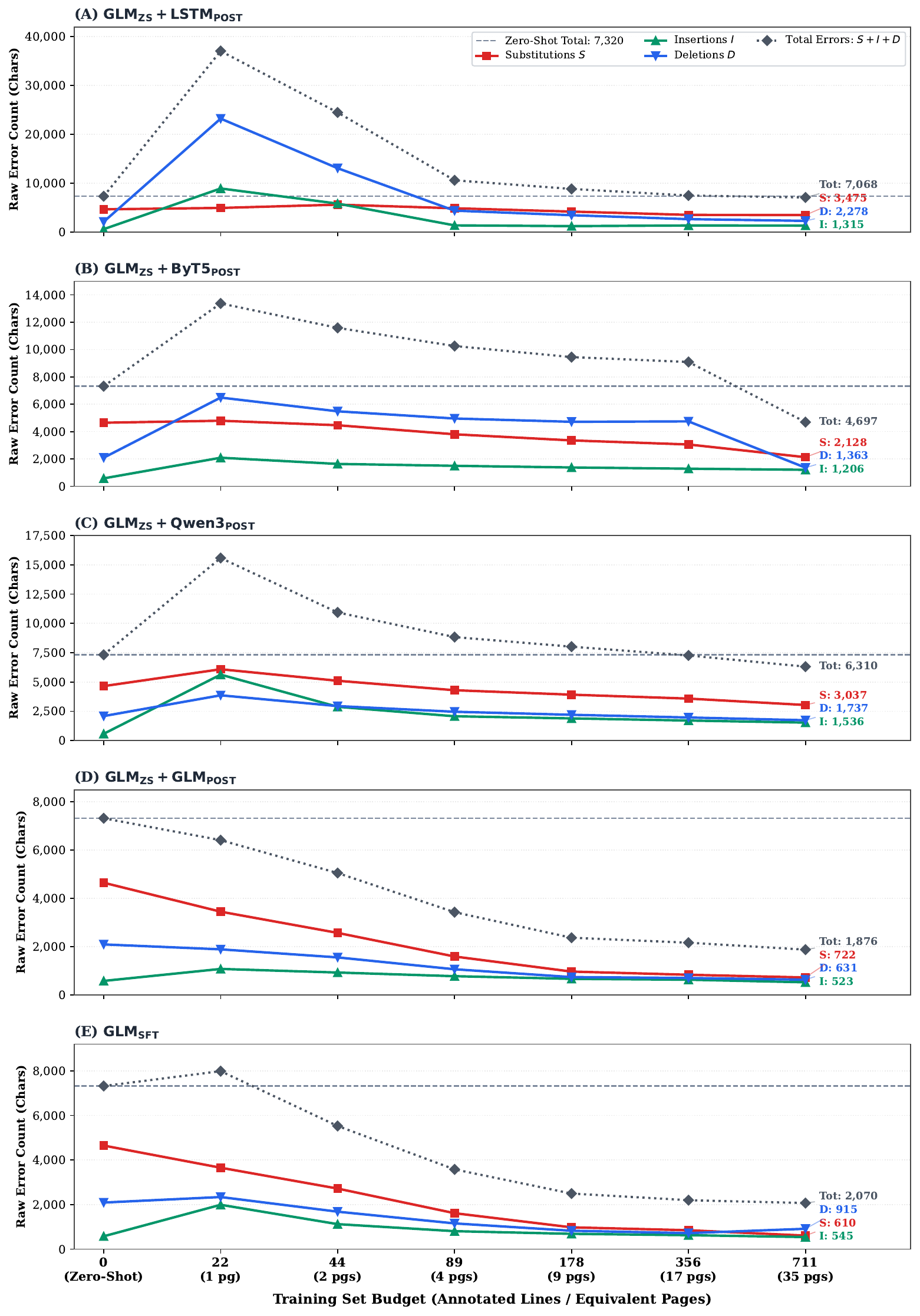}
  \caption{Edit operation counts for each of the training-based methods. For Seq2Seq models, deletions make up the largest proportion of edits. However, for LLM and VLM based methods, substitutions are more common, especially at tiny training budgets. For directly finetuning VLMs for OCR, we noticeably see a jump in insertions at 1 page of training data, likely indicating a spike in hallucinated text.}
  \label{fig:edit_ops}
\end{figure*}